\documentclass [10pt,twocolumn]{IEEEtran}
\usepackage{algorithm,amsbsy,amsmath,amssymb,epsfig,bbm,mathrsfs,multirow,amsthm}
\usepackage[T1]{fontenc}
\usepackage[latin9]{inputenc}
\usepackage{array,multirow}
\usepackage{color}
\usepackage{cite}
\usepackage{float}
\usepackage{makecell}
\usepackage[x11names,dvipsnames,table]{xcolor} %for use in color links
\usepackage{colortbl} 
\usepackage{multirow}
\usepackage{subcaption} 
\usepackage{lipsum}
\usepackage{booktabs}
\definecolor{mygray}{gray}{0.6}
\definecolor{myblue}{rgb}{0.8,0.85,1} 
\usepackage{array}
\newcolumntype{L}[1]{>{\raggedright\let\newline\\\arraybackslash\hspace{0pt}}m{#1}}
\newcolumntype{C}[1]{>{\centering\let\newline\\\arraybackslash\hspace{0pt}}m{#1}}
\newcolumntype{R}[1]{>{\raggedleft\let\newline\\\arraybackslash\hspace{0pt}}m{#1}}

\usepackage{array, tabularx, boldline}
\usepackage{eurosym}
\usepackage{amstext} % for \text
\DeclareRobustCommand{\officialeuro}{%
  \ifmmode\expandafter\text\fi
  {\fontencoding{U}\fontfamily{eurosym}\selectfont e}}

\usepackage{epstopdf}
\usepackage{supertabular}
\usepackage[noend]{algpseudocode}  % For pseudocode
\usepackage{mathtools}
\usepackage{url}
\usepackage{authblk}

\usepackage{array, tabularx, boldline}
\usepackage{graphicx}
\usepackage{cellspace}
\newcolumntype{b}{X}
\newcolumntype{s}{>{\hsize=.5\hsize}X}

\title{Predictive Maintenance for Edge-Based Sensor Networks: A Deep Reinforcement Learning Approach}
\author[*]{Kevin Shen Hoong Ong}
\author[*]{Dusit Niyato}
\author[$\dag$]{Chau Yuen}
\affil[*]{School of Computer Science and Engineering, Nanyang Technological University Singapore}
\affil[$\dag$]{Engineering Product Development, Singapore University of Technology and Design}
\begin{document}
%\title{\huge Sensor Networks using Deep Reinforcement Learning} % Smoke-Screen Title
%\title{\huge Predictive Maintenance of Equipment-based Sensor Networks with Deep Reinforcement Learning}
%\author{Kevin Shen Hoong Ong, Dusit Niyato \\
%\IEEEauthorblockA{\textit{School of Computer Science and Engineering} \\
%\textit{Nanyang Technological University Singapore}\\
%ongs0129@ntu.edu.sg, dniyato@ntu.edu.sg}
%}

\maketitle
%====================================================================
\begin{abstract}
Failure of mission-critical equipment interrupts production and results in monetary loss. The risk of unplanned equipment downtime can be minimized through Predictive Maintenance of revenue generating assets to ensure optimal performance and safe operation of equipment. However, the increased sensorization of the equipment generates a data deluge, and existing machine-learning based predictive model alone becomes inadequate for timely equipment condition predictions. In this paper, a model-free Deep Reinforcement Learning algorithm is proposed for predictive equipment maintenance from an equipment-based sensor network context. Within each equipment, a sensor device aggregates raw sensor data, and the equipment health status is analyzed for anomalous events. Unlike traditional black-box regression models, the proposed algorithm self-learns an optimal maintenance policy and provides actionable recommendation for each equipment. Our experimental results demonstrate the potential for broader range of equipment maintenance applications as an automatic learning framework.
%
%{\it Keywords}- Deep reinforcement learning, Deep Q-learning, cognitive radio networks,  wireless, sensor networks.
\end{abstract}
%main
%\vspace{-4mm}
%=====================
\section{Introduction}
%=====================
% main contributions of the paper
%Write something to tell a story what problem am I trying to solve(i.e WHAT), why is it important I solve the problem (WHY - motivation), HOW do I plan to solve the problem (HOW), 

Equipment downtime is generally defined as the outage time that accumulates whenever production process stops and current world class standards for downtime is $\leq$10$\%$~\cite{Christer2019}. Despite the rapid technological advances and increasing equipment complexity, frequent occurrence of equipment downtime remains and often results in monetary loss. Maintenance teams are given limited maintenance budgets and face tremendous cost pressure to ensure that the production line is always operational. In an event of an unplanned equipment fault, the corrective maintenance option is performed to bring the failed equipment up to its operational status to meet the product delivery deadline. Forward looking companies employ preventive maintenance to reduce the likelihood of unplanned equipment downtime through scheduled upkeep of production equipment condition. Although the long-term goal of reducing overall maintenance costs has been touted, manufacturing productivity is slightly improved at the expense of higher maintenance cost. Predictive maintenance is overall considered more efficient (i.e manpower and cost) and equipment downtime is minimized because maintenance is only performed based on the real-time condition of an equipment.  

%Unlike Supervised Learning approach, which requires manual feature-engineering of a domain-specific rule-based policy, Reinforcement Learning (RL) will derive the rules by itself through interaction with the environment. Equipment lifetime can be considered as a series of state transitions from an initial healthy state to a state that is completely unhealthy (i.e failure). Assuming the time to failure behavior is stochastic and fully observable, a reinforcement learning agent can be tasked to observe any sensor device's operation. Once the optimal policy is learned, the agent will be able to provide the necessary maintenance recommendations. In this paper, we model the problem into a Reinforcement Learning framework and propose the use of Deep Q-Learning to solve.  

As industries worldwide journey towards the Industry 4.0 vision to boost manufacturing productivity, modern equipments become increasingly complex and requires longer maintenance time. Admist a manpower lean economy that is driven by productivity, one of the key challenges resides in the latent demand to simplify the complexity of machine sensor data interpretation for predictive maintenance purposes. Traditional black-box regression models are feature-engineered towards domain-specific applications, and solution extensibility to similar applications and feature updates request are costly short-term endeavours. Deep Learning (DL) have recently been proposed as an alternative for estimation of remaining useful life of equipment~\cite{babu2016deep,zheng2017long,jayasinghe2018temporal} and ball bearings~\cite{sutrisno2012estimation, guo2017recurrent}. Similar work has also been reported in~\cite{zhang2018equipment} to learn the health indicators for the remaining useful life estimation of a turbofan engine by using Temporal Difference (TD) learning. However, these approaches are only concerned about accurate estimation of the equipment's remaining operational uptime and lack meaningful insights to support the maintenance team's decision-making process. Common reasons can be attributed to the fact that explicit models of real-world problems are largely unknown or too complex to accurately model with traditional model-based approaches. In this work, we propose a model-free Deep Reinforcement Learning (DRL) algorithm approach to self-learn optimal maintenance decision policies, from the health state of an equipment, more importantly with actionable recommendation. 

Increased pressure on margins, higher customer expectations and the declining cost of cloud computing are enticing factors for manufacturers to stay competitive. Conversely, both manufacturing and equipment-maker companies remain security averse to transmitting sensitive production sensor data to the cloud, via unreliable internet connections, and any adoption is generally reserved only for companies with deep pockets~\cite{liu2017industry}. An alternative approach is to utilize a locally interconnected sensor network of equipment. In our proposed system model, each equipment is equipped with sensor-based edge computing devices~\cite{ong2010towards}, which can share data and communicate with other edge devices or equipment. Overall benefits of adopting this approach includes time-efficient decision-making process, a responsive yet data informed maintenance support team and reduction in network traffic across the factory shopfloor IT infrastructure.

In summary, the contributions of this paper are as follows:
\begin{enumerate}
	\item Problem formulation of maximizing equipment run-time as a function of multiple sensor data input. The decision-policy is obtained by using the model-free DRL approach. 
	\item The DRL algorithm offers recommendation support for the replacement policy of an equipment with easy-to-understand data-driven recommendations via an Equipment Health Indicator status. %Recommendations are derived from equipment health indicator data, with tangible yet explainable recommendations.
	\item Despite random initial health states and an absent ground truth, the proposed DRL algorithm tackles the sparsely-dense reward maintenance problem with almost consistent recommendations across similar datasets and equipment.
	%\item A threshold-based feature has also been introduced with comparable performance to non-threshold based technique. With this feature, the paper hopes the approach is more amenable to industrial practitioners.
	%\item Modifications to proposed DRL method has shown prediction performance improvements that is as good as state-of-the-art machine learning algorithms, used in equipment health monitoring. \textbf{(To be Validated)}
\end{enumerate}

\vspace{-2.6mm}
%==================
\section{System Model}
\label{sec:sys_model}
\begin{figure}[htbp]
    \centering
    \includegraphics[width=0.32\textwidth]{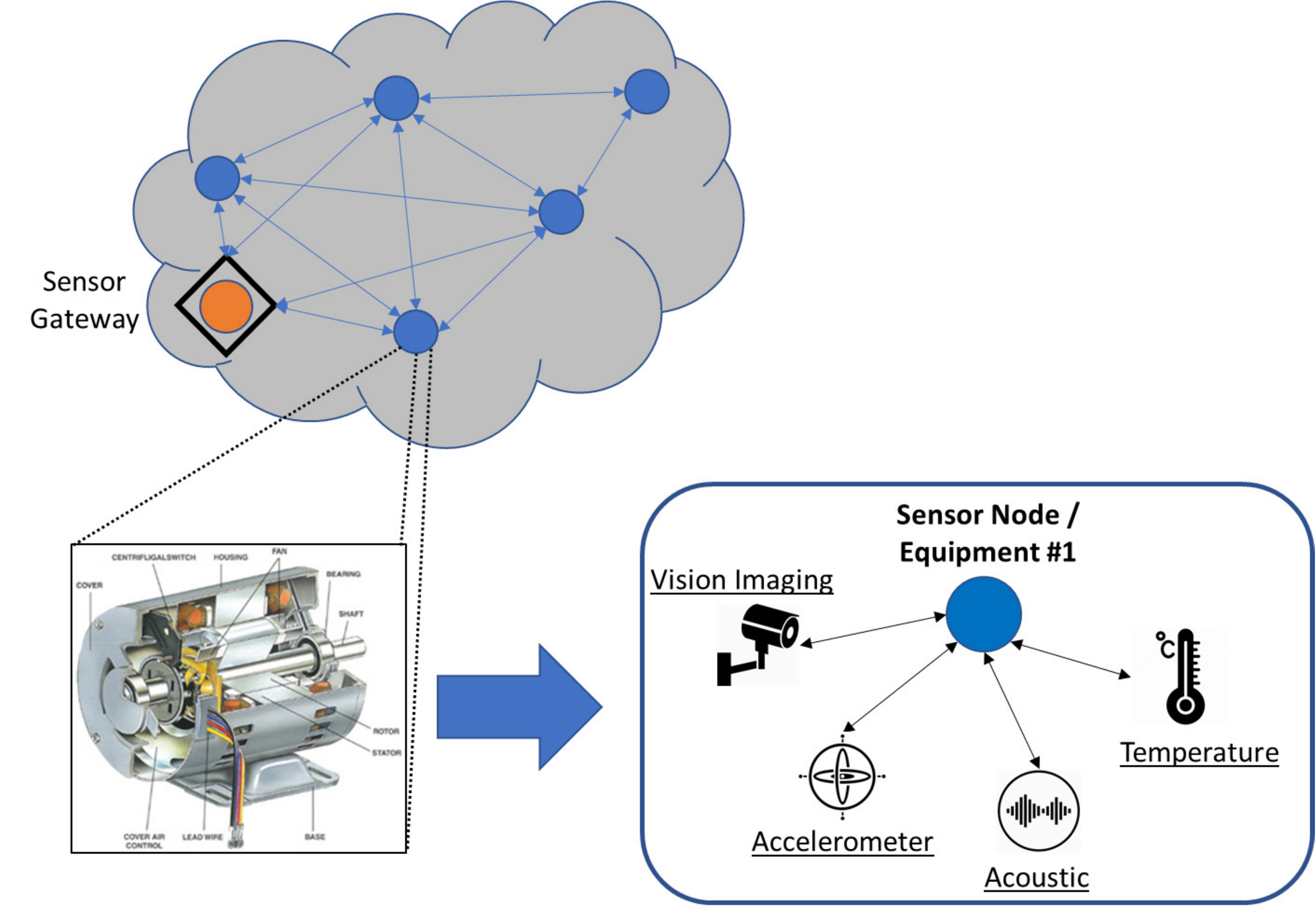}
	\caption{System Model of Sensor Network at Equipment Level} 
	\label{fig:system_model}
\end{figure}

%==================
%To construct the model, is the transition known or unknown? To take an action, in a state, is it stochastically or deterministically for a agent to transition to new state? Same applies for policy (i.e deterministic or Random).
Consider the application of a sensor network for the purpose of equipment health monitoring, either through retrofitting or in-situ configuration. Our proposed system model resembles a Star Topology network at the Equipment level, see Figure~\ref{fig:system_model}. From the current configuration, the aggregated data can be propagated to a larger Sensor Network, such as tree or mesh topology configurations, for more complex data analysis. Typical network components include Sensor Nodes (SN) and a Sensor Gateway (SG), also known as Base Station. Please note that the terms Base Station and SG are used interchangeably. 

SN devices are model representations of physical sensors transmitting data such as temperature, accelerometer, gyroscope and acoustic, depicted in Figure~\ref{fig:system_model}. Concurrent streaming of raw sensor data is known to consume high network bandwidth and degrades overall network performance. To mitigate network congestion scenario, SG-based data aggregation and processing at the SG node is proposed. The SG node is assumed to be always powered with adequate computation and storage capabilities. Next,  sensor data analysis will be performed using batch processing (i.e time-frame comprising of fixed time-step duration) of sensor data for memory and computation efficiency purposes. Without loss of generality and for ease of presentation, a pair of SG and SN is assumed. 
%Existing industrial practice utilizes network traffic prioritization methods, such as Quality-of-Service (QoS) configurations and network segmentation, to minimize the occurrence of network congestion and data loss. 

Data analysis is performed on window size with $\mathcal{K}$ time-steps and the output of the prediction are three maintenance action: \textit{Repair}, \textit{Replace} and \textit{Hold}. \textit{Hold} is the default action where SG predicts a low probability of imminent failure to occur. In the event that an anomalous sensor reading is observed with high certainty, the SG node has to decide which of the two recommendations, \textit{Replace} or \textit{Repair}, should be provided to the maintenance team - mirroring real-life decision-making. In practice, the \textit{Repair} action is almost certainly executed given the busy schedule and maintenance budget constraints. 

%==================
\section{Problem Formulation}
\label{sec:problem_formulation}
%==================

%The environment provides the agent with a suitable observation of the current state (e.g. an image, video, sensory data, etc.) which gets processed by the agent through a policy (e.g. a convolutional neural network) outputting the most likely action in that current state which could then executed by the agent in its environment. The Environment now responses with a reward signal evaluating the quality of that step. It could be a positive reward signal to reinforce certain behavior or a negative one to punish bad decisions. Of course, the whole process is repeated until either the episode terminates by reaching the goal or we reach an upper limit. Some algorithms depend on data collected through this whole episode like Policy Gradient, others just need a batch of {state, action, reward, next state} to be learned.

In our system model, the challenge is to maximize the total run-time of the equipment with maintenance budget constraints. The objective function is described in~(\ref{eqn:reward_function_1}) as Maximum Equipment UpTime ($\rho$) and is cast into a Markov Decision Process (MDP) framework with fully observable states. MDP is formally described as a 5-tuple consisting of state ($\mathcal{S}$), action ($\mathcal{A}$), probabilistic distribution of state transitions ($\mathcal{P}$), reward function ($\mathcal{R}$) and discount factor ($\gamma$). Mathematically, the tuple can be compactly denoted as ($\mathcal{S}$, $\mathcal{A}$, $\mathcal{P}$, $\mathcal{R}$, $\gamma$). 
%\vspace{-0.5mm}
\begin{equation}
   \label{eqn:reward_function_1}
   {\rho} = \sum_{Node=1}^{N}{RunTime}
\end{equation}

Given a sensor network, the generated sensor data is denoted as $x_t^i$, where $i \in \left\{ {0, 1, \ldots, Z} \right\}$ at every time-step ($t$). From (\ref{eqn:reward_function_1}), $N$ represents the number of SN devices within the considered sensor network, and $N=1$ is assumed. The sensor data is then discretized and simplified in~(\ref{eqn:discrete_sensor_data}). Then, we generalize the state space of each sensor node in~(\ref{eqn:sensor_state_space}). 
\vspace{-1mm}
\begin{equation}
	\label{eqn:discrete_sensor_data}
	{q_x} \leftarrow Discretize(x_t^i)	
\end{equation}
\begin{equation}
	\label{eqn:sensor_state_space}
	\mathcal{S}^i = \left\{ {{q_x}} \right\}
\end{equation}

\begin{figure}[htbp]%[!bp]%[!htb]
   \vspace{-6mm}
    \centering
    \includegraphics[trim={60 5 70 40},clip,width=0.38\textwidth]{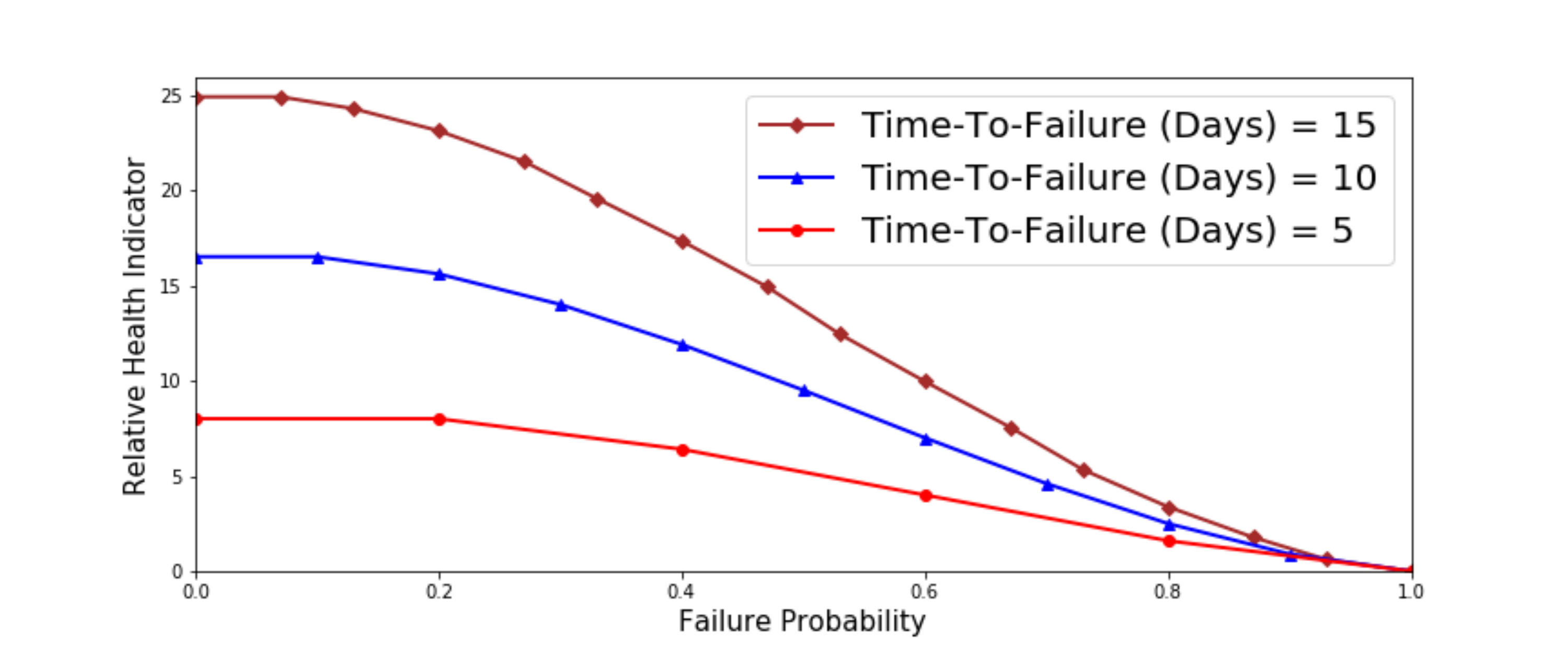}
	\caption{Health Degradation wrt Equipment Failure Probability} 
   %\vspace*{-4mm}	   
	\label{fig:Eqpt_Failure_Prob_TTF}
	\vspace*{-2mm}
\end{figure}

For every equipment or sensor manufactured, the manufacturers specify the Mean-Time-Between-Failure (MTBF) and operating temperature information in the technical datasheet. Due to the elasticity of environmental temperature, the equipment's degradation rate and state change is inadvertently non-sequential. In this work, we consider the sensor state changes to be initially sequential with decreasing trend over finite time-steps. For example, assume the probability of state change increases as operating temperature increases. Eventually, an obvious temperature mode change occurs, and the rate of sensor degradation decays exponentially with respect to time, described in~(\ref{eqn:reliability}). Within the current operating temperature mode, the sensor's state change would skip multiple states to reflect corresponding temperature changes. The observed state change behavior is known to mirror a concave exponential decay trend with respect to equipment run-time. When the transition probability of the state-action pair is considered, a pseudo health indicator is derived and shown to vary according to increasing failure probability, see Figure~\ref{fig:Eqpt_Failure_Prob_TTF}. Notably, the health degradation trend clearly illustrates inverse correlation to the increasing sensor state values and is aligned with the proposed system model.
%\vspace{-0.5mm}
\begin{equation}
\label{eqn:reliability}
	\mathcal{F}(t) = e^{-\lambda t}
\end{equation}

To model the environmental state ($S^{\tau}$), we consider and simplify the operating temperature conditions ($\tau$) into a binary form with conditional constraints, see (\ref{eqn:temp_state_space}). To be clear, units of $\tau$ is in degree Celsius while Low and High operating temperatures are binarily represented. % as $0$ and $1$ respectively. 

\begin{equation}
	\label{eqn:temp_state_space}
	S^{\tau}=\left\{ \begin{array}{l}
			{0, \; \; if \;\tau  \in [25,60]}\\
			{1, \; \; if \;\tau  > 60}		
			\end{array} \right.
\end{equation}

The resultant state space for our system model is summarized as the Cartesian product:
\begin{equation}
	\label{eqn:systemmodel_state_space}
	\mathcal{S} = \mathcal{S}^i \times S^{\tau}
\end{equation}

%Action Space \\
Next, the model's action space is encoded as a vector of scalar actions: \textit{Replace}($\epsilon$), \textit{Repair}($\eta$) and \textit{Hold}($\kappa$). Let us assume that a maintenance account ($\beta$) is credited to the maintenance agent. $\kappa$ represents the agent's default action and the associated maintenance cost is zero. As the equipment state (i.e. sensor health values) starts degrading, the frequency and cost of repairs will gradually increase with time before increasing exponentially. Hence, the agent is tasked to decide the appropriate sequence of actions to perform at each state as described in (\ref{eqn:action_space}). To mimic maintenance decision-making process, the imposed cost constraints ($\mathcal{C}$) ensures that the agent derives a sensible maintenance policy where $\beta$ is not violated within the given time frame.
%\vspace{-2mm}
%~\cite{mathew2003strategy}. Hence, the agent is tasked to decide the appropriate sequence of actions to perform at each state as described in (\ref{eqn:action_space}). To mimic maintenance decision-making process~\cite{ArtTroubleshoot2014}, the imposed cost constraints ($\mathcal{C}$) ensures the agent derives a sensible maintenance policy where $\beta$ is not violated within the given timeframe. 
%\vspace{-2mm}
\begin{equation}
	\label{eqn:action_space}
	\mathcal{A}=\left\{ \begin{array}{l}
			{(\epsilon, \eta, \kappa)\;|} \\
			{\sum\limits_{n = 1}^N {\mathcal{C}_\epsilon} \geq 2\sum\limits_{n = 1}^N {\mathcal{C}_\eta}\; \;{\text{and}}\; \; \beta - \sum\limits_{n = 1}^N {\mathcal{C}_\epsilon} \geq 0,} \\
			{\sum\limits_{n = 1}^N {\mathcal{C}_\epsilon} \leq \sum\limits_{n = 1}^N ({\mathcal{C}_\eta}/2) \; \; {\text{and}}\; \; \beta - \sum\limits_{n = 1}^N {\mathcal{C}_\eta} \geq 0}
			\end{array} \right.
\end{equation}

%We consider \textit{Repair} cost ($\epsilon$) as the sum total of cost involving component replacement and manpower cost is initially assumed to be less than half the cost of performing a \textit{Replace} ($\eta$) and the agent will perform the \textit{Repair} action~\cite{ArtTroubleshoot2014}. Otherwise, \textit{Replace} or \textit action is performed when . In the end,  For easy reference, the depicted scenarios are described in Equation~\ref{eqn:action_space}. %Just to be clear, the values of $\epsilon_{Repair}$ and $\epsilon_{Replace}$ are user-defined and details are mentioned in Section~\ref{subsec:param_config}. For easy reference, the depicted scenarios are described in Equation~\ref{eqn:action_space}. 

%\begin{figure}%[!htb]
%    \centering
%    \includegraphics[width=0.35\textwidth]{Figure/temperature_state}
%	\caption{Temperature State Space and Transitions}    
%	\label{fig:state_change_temperature}
%	%\vspace*{-3mm}
%\end{figure}

\begin{figure}[htp]%[!htb]
   \vspace{-6mm}
    \centering
    \includegraphics[width=0.4\textwidth]{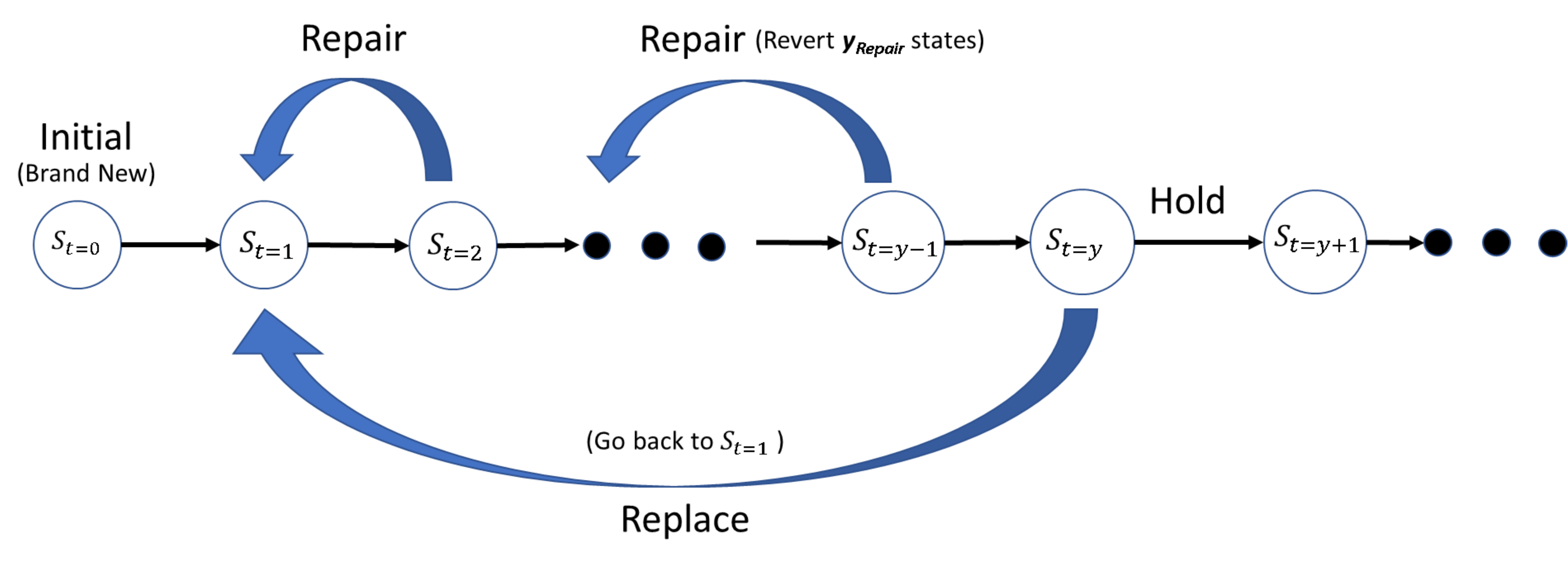}
	\caption{Example State Space and Transition with Replace and Repair Action} 
   %\vspace*{-4mm}	   
		\label{fig:state_change_repair_replace}
		%\vspace{-7mm}
\end{figure}

For model simplification purposes, consider a single equipment with $N=1$ sensor attached. Given observable state changes and transitions, the agent can select random actions to perform. $\epsilon$ resets the sensor's operational state to an almost new condition while $\eta$ seemingly reverts the sensor's operational state by $y_{Repair}$ states, to a previously observed sensor state. In addition, we consider that not all repairs are identical and the sensor state change $\phi(\mathcal{S})$, will vary according to the RepairType($\psi$), where $\psi \in \left\{ {0,1, ..., M} \right\}$. From Figure~\ref{fig:state_change_repair_replace}, $\eta$ is invoked at $\mathcal{S}_t=y-1$ and the selected type of repair invokes a state transition from $\mathcal{S}_t=y$ to $\mathcal{S'}_t=y-|\phi(\mathcal{S})|$. A compact representation of the behaviour is described in (\ref{eqn:repair}).
%\vspace{-1mm}
\begin{equation}
\label{eqn:repair}
	y_{Repair} \in \left\{ {\phi (\mathcal{S})|\mathcal{S} \in \psi} \right\}
\end{equation}

%Given multiple factors, such as varying sensor parts, complexity of state transitions for any sensor and the potentially large state space, Deep Reinforcement Learning (DRL) is proposed. Specifically, the DRL agent will learn the transition probability values through interaction with the environment (eg. training dataset). 

The Reward function is formally described as $\mathcal{R}(s_t, a_t, s_{t+1})$. From the previous derivations, $\mathcal{R}$ can be re-interpreted as the cumulative sum of the sensor runtime with respect to state $\mathcal{S}$ and action $\mathcal{A}$. An immediate reward function $\mathcal{R}_t \in \mathcal{R}$ is proposed in Equation~\ref{eqn:reward_space} to guide the agent's actions: 

\begin{equation}\label{eqn:reward_space}
	\mathcal{R}_t=\left\{ \begin{aligned}
			\mathcal{R}_{\text{Rpl}},&\quad {\text{if}} \; \;\mathcal{S}^{i}_{t} > 0,\; \; \beta > 0 \\
			\mathcal{R}_{\text{Rpa}}, &\quad {\text{if}} \; \;\mathcal{S}^{i}_{t} > 0,\; \; \beta > 0 \\
			\mathcal{R}_{\text{Exp}}, &\quad {\text{if}} \; \;\mathcal{S}^{i}_{t} > 0\; \\
			\mathcal{R}_{\text{Frug}}, &\quad {\text{if}} \; \;\mathcal{S}^{i}_{t} > 0,\; \; \beta > 0 \\
			\mathcal{R}_{\text{Pen}}, &\quad {\text{otherwise}}					
			\end{aligned}\right.
\end{equation}

For example, executing either $y_{\text{Replace}}$ or $y_{\text{Repair}}$ action, within the $\beta$ constraints, will offer the agent an arbitrary high health points for performing Replace ($\mathcal{R}_{\text{Rpl}}$) and Repair ($\mathcal{R}_{Rpa}$) actions respectively. Conversely, the agent is heavily penalized if either $\kappa$ is performed throughout the episode or when the sensor state is zero ($\mathcal{R}_{\text{Pen}}$). Real-world applications have sparse rewards and motivating the agent to explore is very challenging. In this paper, $\mathcal{R}_{\text{Exp}}$ value is arbitrarily defined to intrinsically motivate the agent to explore the environment. To increase the probability of sparse state visitation, where potentially high rewards are found, a proposed ranking mechanism is proposed and described in Section~\ref{ssec:PER}. $\mathcal{R}_{\text{Frug}}$ reward is also defined to encourage the agent to execute the corresponding action optimally, mimicking human decision-making.

Next, the state-value function ($V^{\pi}(s)$) is expressed as:
\begin{equation}
	\label{eqn:state_value}
	V^{\pi}(s)=\mathbb{E}_{\pi}[\sum_{t=0}^{\infty}\gamma\mathcal{R}_{t+1}|S_{t}=s] 
\end{equation}

Then, we utilize the standard RL framework to obtain the optimal policy $(\pi^{*})$. The Markov property is applied on (\ref{eqn:state_value}) and the value function is simplified and re-expressed as:
\vspace{-0.5mm}
\begin{equation}
	\label{eqn:value_func}
	V^{\pi^{*}}(s)=\sum_{a}\pi(s,a)\sum_{s' \in S}\mathcal{P}_{\pi(s)}(s, s')\left[ {\mathcal{R}(s, a)+\gamma V^{\pi}(s')} \right]
\end{equation}

The associated policy function obtains the maximum action that is possible from Equation~\ref{eqn:value_func}. Hence, the Q-function $\mathcal{Q}^{*}(s, a)$ can be updated using the Bellman equation and expressed as:
\begin{equation*}
   \begin{multlined}
      \overbrace{\mathcal{Q}^{*}(s, a)}^{\text{New Q value}}= (1-\alpha)\overbrace{\mathcal{Q}(s, a)}^{\text{Current Q value}}+\alpha[\overbrace{\mathcal{R}(s, a)}^{\text{Reward received}}\\
+\gamma \underbrace{\max\limits_{a' \in \mathcal{A}}\mathcal{Q}'(s', a')}_{\text{Max(Expected future reward)}}]
   \end{multlined}
\end{equation*}
\noindent
where $\alpha$ denotes the learning rate of the agent; $\gamma  \in \left[ {0,1} \right]$ is the discount factor where the agent performs tradeoff between the observed immediate and potential future reward. The $\max$ operator selects the highest-valued state-action pair that  consequently assists in the derivation of the optimal policy. 

Many real-world problems have very large state space, which makes it infeasible to compute and learn all exact state-action transition values. Instead, an approximated estimate of the state-action value pair can be learned using Temporal Difference (TD) learning. As TD learning process resembles a stochastic gradient descent, the updated Q-value is denoted as $\mathcal{Q}\left( {S_t^i,{A_t};{\boldsymbol{\theta _t}}} \right)$ towards a target value of $y_{t}^{Q}$ in (\ref{eqn:Q_func_SGD}).
%\vspace{-2mm}
\begin{equation}
\label{eqn:Q_func_SGD}
y_t^Q \equiv {\mathcal{R}_{t + 1}} + \gamma \mathop {\max }\limits_A \mathcal{Q}\left( {S_{t + 1}^i,\mathcal{A};{\boldsymbol{\theta _t}}} \right)
\end{equation}

In coupling a multi-layered neural network, a Deep Q-Network (DQN), with experience replay and target network ($y_{t}^{DQN}$), performance of the algorithm is vastly improved and achieved fairly good generalization performance~\cite{mnih2013playing}. The target network is a copy of the original Q-network and parameter synchronization occurs every $\tau$ steps, such that $\boldsymbol{\theta _{t}^{-}} = \boldsymbol{\theta _{t}}$, and is represented in~(\ref{eqn:DQN}).
%\vspace{-1mm}
\begin{equation}
\label{eqn:DQN}
y_t^{DQN} \equiv {\mathcal{R}_{t + 1}} + \gamma \mathop {\max }\limits_A \mathcal{Q}\left( {S_{t + 1}^i,\mathcal{A};\boldsymbol{\theta _t^ -} } \right)
\end{equation}

%By referencing the probability transition values, stored in the Q-table, the agent performs an action. These Q-values are iteratively updated for the states the agent has previously visited. Given sufficient exploration time, the agent can construct a policy where the cumulative rewards is maximized, also known as optimal policy. At the same time, the application of Q-learning in the real-world is both limited and infeasible. Common reasons include poor search performance in high-dimensional order of observable states and its inability to estimate values for unseen states. Considering the varying sensor parts, complexity of state transitions for any sensor and the potentially large state space, Deep Reinforcement Learning (DRL) is proposed for Q-function estimation and the DRL agent will learn the transition probability values through interaction with the environment (eg. training dataset). The proposed DRL algorithm is further described in Section~\ref{sec:Intro_DRL}

% ================================
\section{Solution of Deep Q Learning}
\label{sec:Intro_DRL}
% ================================
\subsection{Double Deep Q Learning}
DQN's inherent tendency for value overestimation and biased estimates are caused by random environment noise and $\arg\max$ operator respectively. Double Deep Q-Learning (DDQN) was then proposed~\cite{van2016deep} to stabilize vanilla DQN algorithm by decoupling the choice and evaluation of best action on two separate networks. In this work, the DDQN inputs contain a tuple of sensor data ($x_t^i$) at every time-step and random action is performed by the DDQN agent. The DDQN's target network generates an output of Q-values and is denoted as: 
%\vspace{-0.05mm}
\begin{equation}
y_t^{DoubleDQN} \equiv {R_{t + 1}} + \gamma \mathcal{Q}\left( {{s_{t + 1}},{a^*};\boldsymbol{\theta _t^ -} } \right)
\end{equation}
\noindent
where ${a^*}=\arg \max \mathcal{Q}\left( {{s_{t + 1}},\mathcal{A};{\boldsymbol{\theta _t}}} \right)$. The discounted \textit{$\mathcal{Q}$-value} ($y_t^{DoubleDQN}$) is taken from the target network with weights $\boldsymbol{\theta _t^{-}}$ and the target network weights ($\boldsymbol{\theta _t^{-}}$) are periodically copied from the \textit{$\mathcal{Q}$-value} network. In particular, the agent's performed actions are extracted from the primary \textit{$\mathcal{Q}$-value} network and the future reward evaluation step is taken from the \textit{$\mathcal{Q}$-target} network. 

\vspace{-3mm}
\subsection{Prioritized Experience Replay}
\label{ssec:PER}
The DRL agent is expected to infer an optimal point from the system model or dataset from the reward constraints. With reference to Figure~\ref{fig:Eqpt_Failure_Prob_TTF}, the equipment replacement point is likely to occur further into the equipment operational cycle as failure probability increases exponentially. Likewise within the DRL context, the sparsely-dense-like reward configuration and the likelihood of performing either $\mathcal{R}_{\text{Rpl}}$ or $\mathcal{R}_{Rpa}$, is higher towards the end of the equipment's operational cycle. The gained intuition suggests a normalized equipment health state value of $0.2$ to be the target optimal value. %For fairly critical machines, several machine maintenance personnel we spoke with considers it reasonable estimate.

Although the traditional combination of $\epsilon$-greedy algorithm with Experience Replay (ER) Buffer technique is well-known, the ER Buffer has a bias to repeatedly sample the same highly-rewarding experiences and is ill-posed for a sparse-reward problem, typical of real-world applications. Moreover, the decay rate of $\epsilon$ value over the training set is a hyperparameter and an inefficient approach given varying Time-to-Failure cycles for each equipment. For the mentioned problems, \textit{Prioritized Experience Replay} (PER)~\cite{schaul2015prioritized} is the proposed component to compliment DDQN.

PER is considered to be an enhancement over ER, and it prioritizes the experiences which offers large differences between prediction and the Temporal Difference (TD) target value. Implicitly, the DRL agent can use the TD error magnitude as an indicator to better focus its attention on the least visited states which could yield potentially larger rewards. The magnitude of the TD error ($\left| \delta_i \right|$) can be incorporated into the ER buffer sample as a tuple $\left( s_t, a_t, r_t, s_{t+1}, \left| \delta_i \right|\right)$. To minimize over-fitting problem in the ER method, stochastic prioritization~\cite{schaul2015prioritized} is introduced to generate a probability $P\left( i \right)$, of a selected state-action pair, from the replay buffer, see~(\ref{eqn:PER_stochasticPior}). $p_i$ denotes the priority value of the $i$-th sample in the buffer and $\alpha$ is a hyperparameter that is used to induce randomness within the experience selection for the replay buffer. Pure uniform randomness is denoted by $\alpha=0$ while $\alpha=1$ emphasizes experiences with the highest priorities; $\sum_k$ denotes the normalization of all priority values within the Replay Buffer.
\vspace{-1mm}
\begin{equation}
\label{eqn:PER_stochasticPior}
	P\left( i \right) = \frac{{{p_i}^\alpha}}{{\sum_k {p_k^{\alpha}} }}
\end{equation}
\vspace{-1.5mm}
\begin{equation}
\label{eqn:IS_weights}
{\left( {\frac{1}{N} \cdot \frac{1}{{P\left( i \right)}}} \right)^b}
\end{equation}

During training of the neural network, the experiences sampled must match the underlying distribution with priority sampling, resulting in a bias towards high-priority samples. The weights of frequently seen samples, within the replay buffer, are then adjusted using Importance Sampling Weights (IS) with the effect of reducing bias. From~(\ref{eqn:IS_weights}), $N$ denotes the Replay Buffer size; $P\left( i \right)$ denotes the sampling probability from~(\ref{eqn:PER_stochasticPior}); $b$ is considered a weighting factor to control the degree in which IS affects the learning process and is annealed up to 1 over the duration of training phase. Readers may refer to~\cite{schaul2015prioritized} for additional details on PER. %These weights are more important towards the end of learning when the Q values start to converge. Readers may refer to~\cite{schaul2015prioritized} for additional details on PER.

\vspace{-3mm}
\subsection{Parameter Noise}
Exploration inefficiency of Reinforcement Learning (RL) is a well-known problem and becomes more challenging when applied on a sparse reward problem. Existing RL approach influences the action space policy at each time-step with no influence on the RL agent's decision policy (i.e neural network). Evolution strategies seek to manipulate the decision policy parameters during each rollout and no influence is exerted on the action space policy. \textit{Parameter Noise} (PN)~\cite{plappert2017parameter} is a technique which strikes an in-between balance of the aforementioned approaches with encouraging results for both on-policy and off-policy algorithms. To explain, PN randomly alters the parameters of the RL agent's decisions which introduces a more consistent exploration behaviour and a less confused RL agent. In this work, PN is introduced to improve exploration efficiency for the RL agents for the sparsely-dense reward problem at hand and the pseudocode for the proposed algorithm, Prioritized DDQN-PN Parameter Noise (PDDQN-PN) algorithm is described in Algorithm~\ref{algo:PDoubleDQN_PN}.
%\vspace{-2mm}
\begin{algorithm}[htbp]%[!htbp]%[H]
\caption{Prioritized Double Deep Q-Learning with Parameter Noise (PDDQN-PN)}
%\algsetup{linenosize=\tiny}
\begin{algorithmic}[1]
\State \textbf{Input:} Action space $\mathcal{A}$, mini-batch size $L_{b}$, target network replacement frequency $L^{-}$
\State \textbf{Output:} Optimal policy $\pi^{*}$

\State \textbf{Initialize:} Prioritized Experience Replay Memory $\mathcal{D}_{priority}$ to capacity $N$, Primary network $\mathcal{Q}_{\theta}$, target network $\mathcal{Q}_{\theta^{'}}$, action-value function $\mathcal{Q}$ with random weights, target action-value function $\hat{\mathcal{Q}}$ with weights $\mathcal{\theta^{-}}$=$\mathcal{\theta}$
\For{Episode=1 to $E$}
	\For{timestep=1 to T}
		\State Observe state $s_{t}$ and select $a_{t}\sim\pi \left( {a,s} \right)$
		\State    With probability $\mathcal{\epsilon}$, perform random action
		\State    Otherwise, choose $a_{t}=argmax_{a}\mathcal{Q}(s_{t}, a)$ from $\mathcal{Q}(s,a;\theta)$;
		\State Execute $a_{t}$ and  receive reward $r_{t}$
		\State Observe next state $s'$ 
		\State Store tuple $(s_{t}, a_{t}, r_{t}, s')$ in $D$ with max priority
		\State Sample random mini-batch transitions, size $(L_{b})$, from $\mathcal{D}_{priority}$, according to transition priority
		\State $y_{t}^{DDQN}=\left\{ \begin{array}{l}
			r, \text{if episode terminates at timestep+1} \\
			r+\gamma max_{a'}\hat{\mathcal{Q}}(\mathcal{\phi}_{j+1}, a', \mathcal{\theta}^{-}),\text{else}
			\end{array} \right.$
		\State Perform Gradient Descent on $(y_{t}^{DQN}-\mathcal{Q}(s_t, a_t))^{2}$ and update priority of each transition
		\State Reset $\theta^{-}$=$\theta$ every $L^{-}$ steps
		\State Update $s_{t} \leftarrow s'$ 
		\State Increment timestep by 1
	\EndFor \textbf{repeat until} timestep is > T, terminate
\EndFor \textbf{repeat until} Episode is > $E$, terminate
\end{algorithmic}
\label{algo:PDoubleDQN_PN}
\end{algorithm}

%=====================
\section{Experiment Study}
\label{sec:experiment_study}
%=====================
%\vspace{-2mm}
\subsection{Dataset}
The NASA Commercial Modular Aero-Propulsion System Simulation dataset (C-MAPSS)~\cite{saxena2008damage} is widely used in literature and is selected to test our model. The data set is generated from Turbofan Engine Degradation Simulator and contains measurements which mimics the degradation behaviour of multiple turbofan engines under various operating conditions. The corresponding fault conditions and complex relations with multiple sensor measurements are listed within four similar smaller datasets (FD001\textasciitilde{}FD004). For simplicity, engine datasets FD001 and FD003 were selected and brief dataset information is shown in Table~\ref{tab:cmapss_dataset_summary}. Within each dataset, 26 columns of data are present. Columns 1 and 2 refers to engine unit and particular engine cycle; Columns 3 to 5 are operating conditions, such as temperature; Columns 6 to 26 contains 21 raw sensor readings. 

\begin{table}[htbp]%[!htbp]
	\centering
	\begin{tabular}{ccc} \hline 
		\textbf{Dataset} & \textbf{FD001} & \textbf{FD003} \\ \hline
		Training Trajectories & 100 & 100 \\ \hline
		Testing Trajectories & 100 & 100 \\ \hline
		Operating Conditions & 1 & 1 \\ \hline		
		Fault Conditions & 1 & 2 \\ \hline				
	\end{tabular}
	\caption{C-MAPSS DataSet~\cite{saxena2008damage} under test}  
	\label{tab:cmapss_dataset_summary}
	\vspace{-6mm}
\end{table} 

%\vspace{-3mm}
\subsection{Data Preparation}
Individual sensor values have been normalized for each sensor type (i.e sensor data column):
%\vspace{-2mm}
\begin{equation}
	\hat x = \frac{{\left({x_i - \mu_i} \right)}}{\sigma}
\end{equation}
\noindent
where $\mu_i$ and $\sigma$ denotes the mean value and standard deviation for the sensor types respectively. $x_i$ denotes the $i$-th sensor value within the corresponding sensor type dataset.

Remaining Useful Life (RUL) calculation is then appended to the normalized data and linear regression is performed. Once the sensor parameters are identified, an RUL distribution graph is plotted and a Gaussian distribution was observed for all data. Sensor data is then dimensionally reduced using principal component analysis and the equipment health indicator is obtained. In the absence of ground truth for equipment health values,~\cite{saxena2008damage} suggested that the degradation behaviour should resemble an exponential decay function like model: 
\begin{equation}
	H(t) = 1 - d - {e^{\{ a{t^b}\} }}
\end{equation}
\noindent
where $d$ denotes the non-zero initial degradation, and $a$ and $b$ are weighted coefficients. Similar degradation behaviour was observed for turbofan equipment health indicator dataset, see Figure~\ref{fig:prediction_results}. Note that in the absence of a ground truth for both repair and replacement action, only the replacement action is tested and a medium-criticality equipment is assumed.

% Note to Kevin: No visible difference between both processed and unprocessed data. Actual reference was ommitted due to rendering problems.
%\begin{figure}
%	\centering
%	%\vspace{-3mm}	
%	\begin{subfigure}[t]{0.42\linewidth}
%		\centering
%		\includegraphics[width=1\linewidth]{Figure/fd001_76}
%		\caption{\label{fig:health_degradation_1}Engine76 - (FD001)[Proc.]}
%	\end{subfigure}
%	~
%%	\begin{subfigure}[t]{0.43\linewidth}
%%		\centering
%%		\includegraphics[width=1\linewidth]{Figure/fd003_69}
%%		\caption{\label{fig:health_degradation_2}Engine69 - (FD003)[Proc.]}
%%	\end{subfigure}
%%	~
%	\begin{subfigure}[t]{0.42\linewidth}
%		\centering
%		%\includegraphics[trim={90 7 90 7},clip,width=1\linewidth]{Figure/fd001_76}
%		\includegraphics[width=1\linewidth]{Figure/fd001_76}		
%		\caption{\label{fig:health_degradation_3}Engine76 - (FD001)[Raw]}
%	\end{subfigure}	
%%	~
%%	\begin{subfigure}[t]{0.43\linewidth}
%%		\centering
%%		\includegraphics[width=1\linewidth]{Figure/fd003_69}
%%		\caption{\label{fig:health_degradation_4}Engine69 - (FD003)[Raw]}
%%	\end{subfigure}	
%	\caption{Example: Health Indicator Degradation Trend}
%	\label{fig:health_degradation}
%	\vspace{-4mm}	
%\end{figure}

\vspace{-3mm}
%=====================
\subsection{Results}
%=====================
PDDQN-PN was implemented using Tensorflow based deep learning library~\cite{stable-baselines} and Adam optimizer was selected. 5000 time-steps of warm-up was performed to fill the agent's memory, using random policy, before commencing actual training. To balance between exploration and exploitation, the exploration fraction was set to 80$\%$ of the number of training simulation time-steps. %For example, given $1.5\times10^4$ time-steps to train the DRL algorithm, data exploration gradually decreases to zero before exploiting the PER buffer from $1.2\times10^4$ timestep.  
\vspace{-3mm}
\begin{figure}[!htbp]
	\centering
    \includegraphics[trim={50 30 100 80}, clip, width=0.36\textwidth]{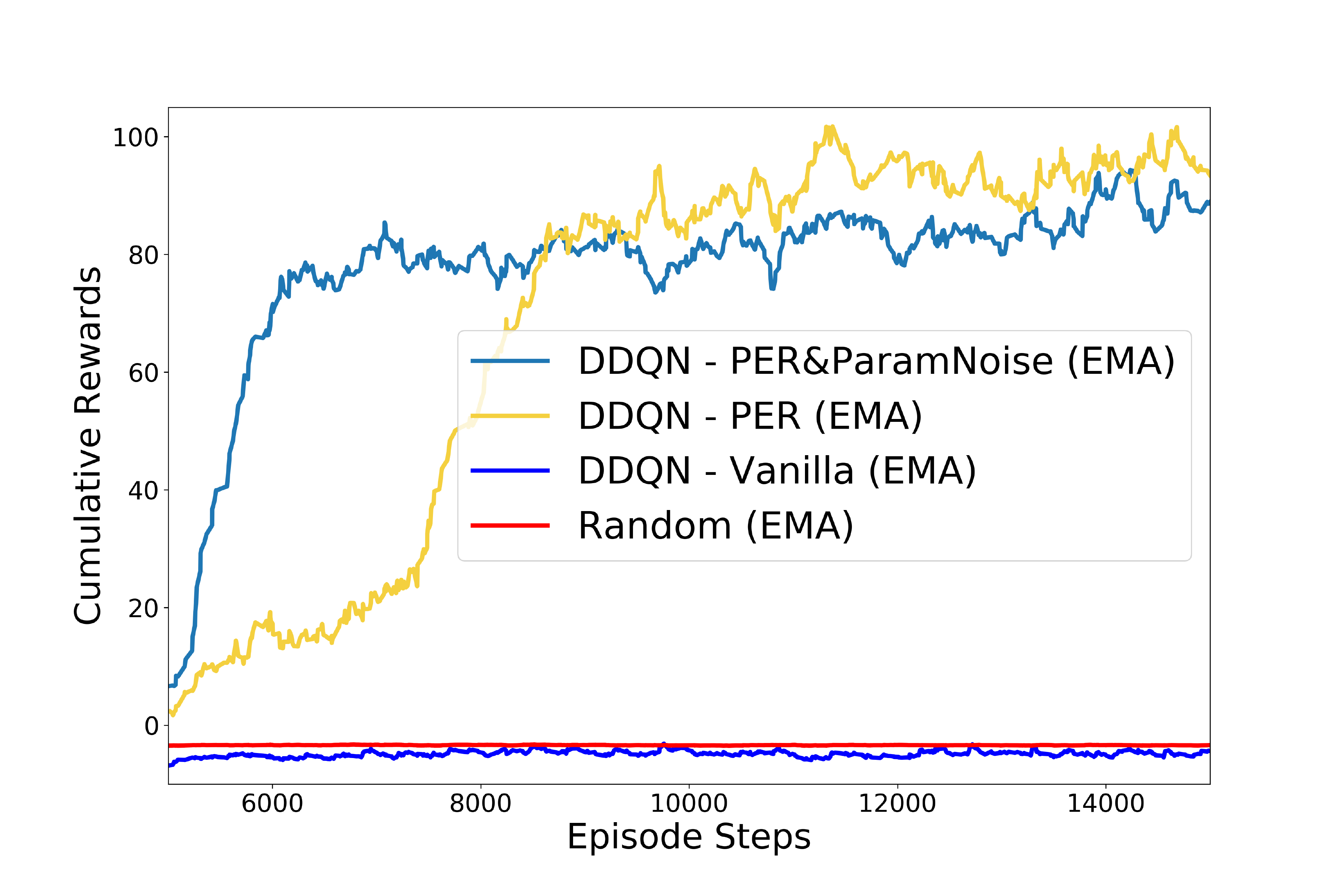}
	\caption{Learning Performance Benchmark between proposed components and Random policy}    
	\vspace{-2mm}	
	\label{fig:performance_results}
\end{figure}

In order to demonstrate the benefits of the proposed algorithm, separate tests were conducted on engine \#76 sensor data from FD001. As a result, the learning performance contribution factor of each sub-component is clearly illustrated. PER component significantly contributed to the DRL agent's ability to learn an optimal policy within $1.2\times10^4$ time-steps and a mean cumulative reward score of 95. Once stacked with parameter action noise, a notable $50\%$ improvement in learning efficiency is clearly observed with a performance loss of $10\%$. From Figure~\ref{fig:performance_results}, it is likely that the mean score for both DDQN-PER and PDDQN-PN will converge given longer simulation time-steps and reduction in exploration time. DDQN-Vanilla performed marginally worse than a random policy given the sparse-reward problem and the results are illustrated in Figure~\ref{fig:performance_results}. For improved readability, Exponential Mean Average (EMA) smoothing factor of $0.18$ was performed on the reported results.

The model prediction method is used to quantify and analyze the DRL agent's learnt decision policy. The prediction results in Table~\ref{tab:prediction_result} noted identical median for DDQN based algorithms with PDDQN-PN reporting the largest standard deviation of $1.27\times10^{-2}$. The higher deviation is attributed to the action parameter noise technique. The random policy agent clearly demonstrated its inability to learn a suitable policy and DDQN-Vanilla offers conversely comparable prediction performance to the DDQN-PER variant. 

\begin{table}[htbp]%[!htbp]
	\centering
	\begin{tabular}{ccc} \hline 
		\textbf{Algorithm} & \textbf{Median} & \textbf{Standard Deviation} \\ \hline
		DDQN - PER + ParamNoise & 0.170 & 0.013 \\ \hline
		DDQN - PER & 0.170 & 0.011 \\ \hline
		DDQN - Vanilla & 0.170 & 0.011 \\ \hline		
		Random & 0.897 & 0.0 \\ \hline				
	\end{tabular}
	\caption{Average Model Prediction Results for Engine \#50}
	\label{tab:prediction_result}
	\vspace{-2mm}
\end{table} 

\begin{figure}[htbp]
	\centering
	\begin{subfigure}[t]{0.45\linewidth}
		\centering
		\includegraphics[trim={20 5 56 40},clip,width=1\linewidth]{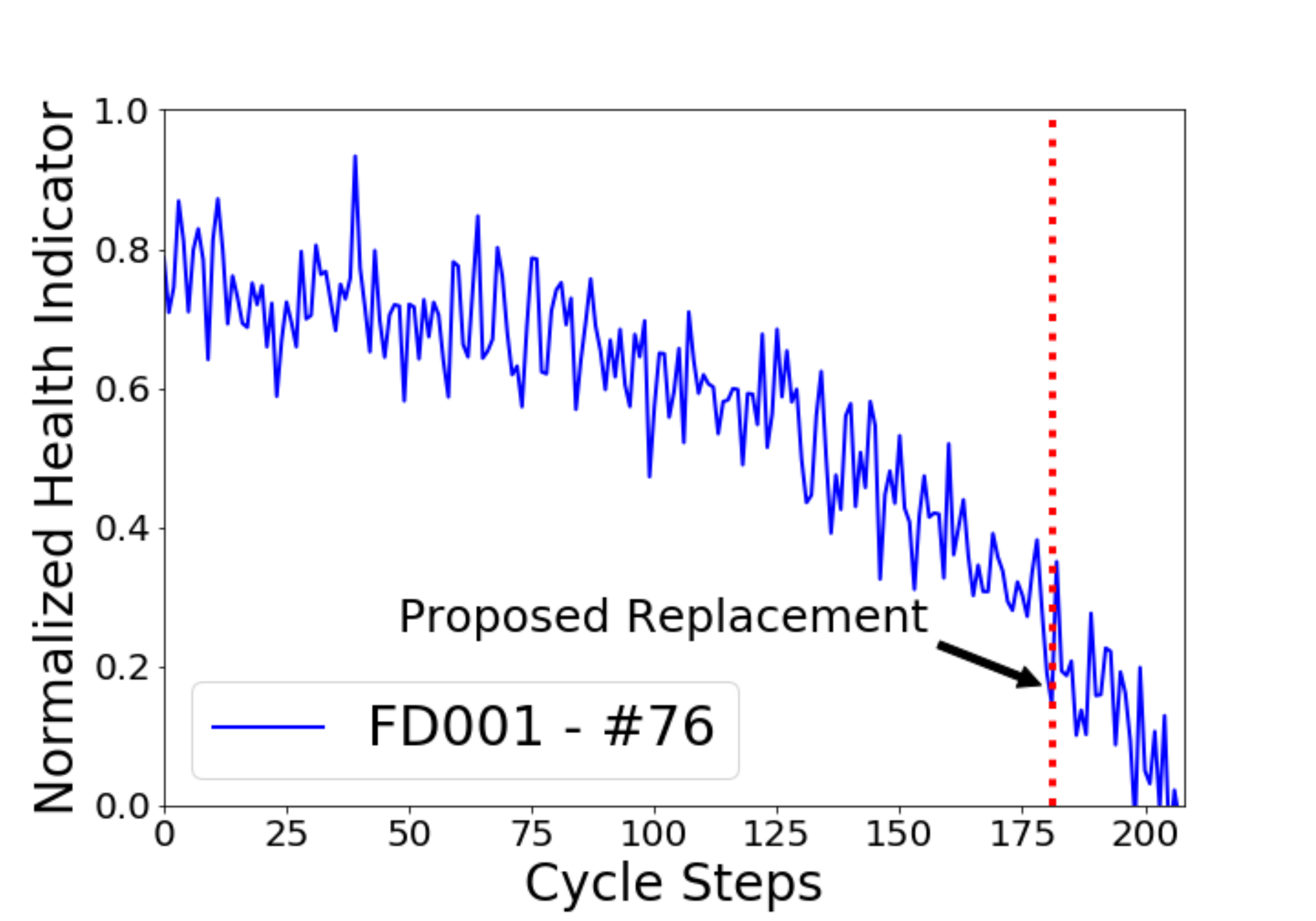}
		\caption{\label{fig:train_fd001_76}Engine76 - (Training)}
	\end{subfigure}
	~
	\begin{subfigure}[t]{0.45\linewidth}
		\centering
		\includegraphics[trim={20 5 56 40},clip,width=1\linewidth]{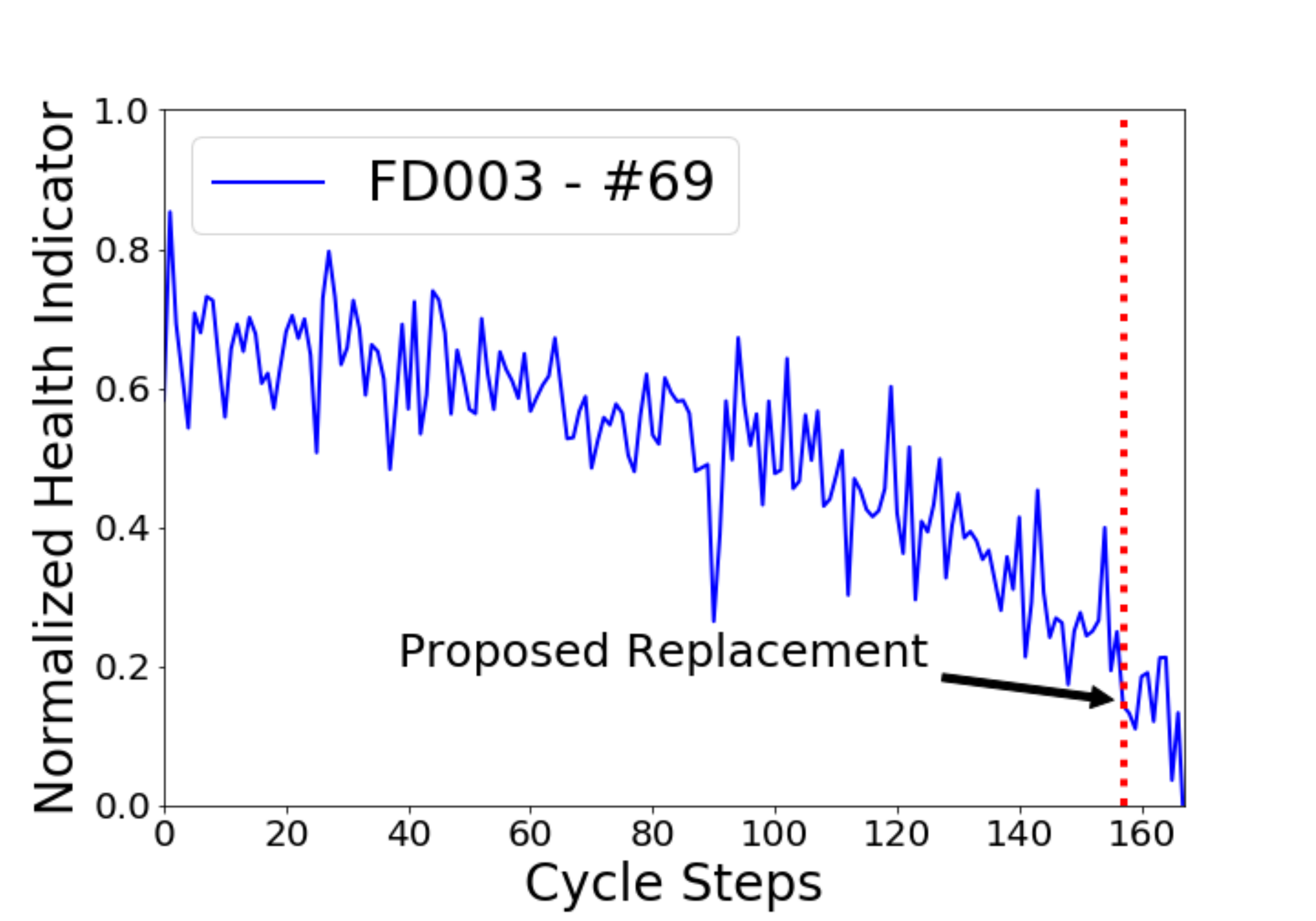}
		\caption{\label{fig:train_fd003_69}Engine69 - (Training)}
	\end{subfigure}	
	~
	\begin{subfigure}[t]{0.45\linewidth}
		\centering
		\includegraphics[trim={-10 5 56 40},clip,width=1\linewidth]{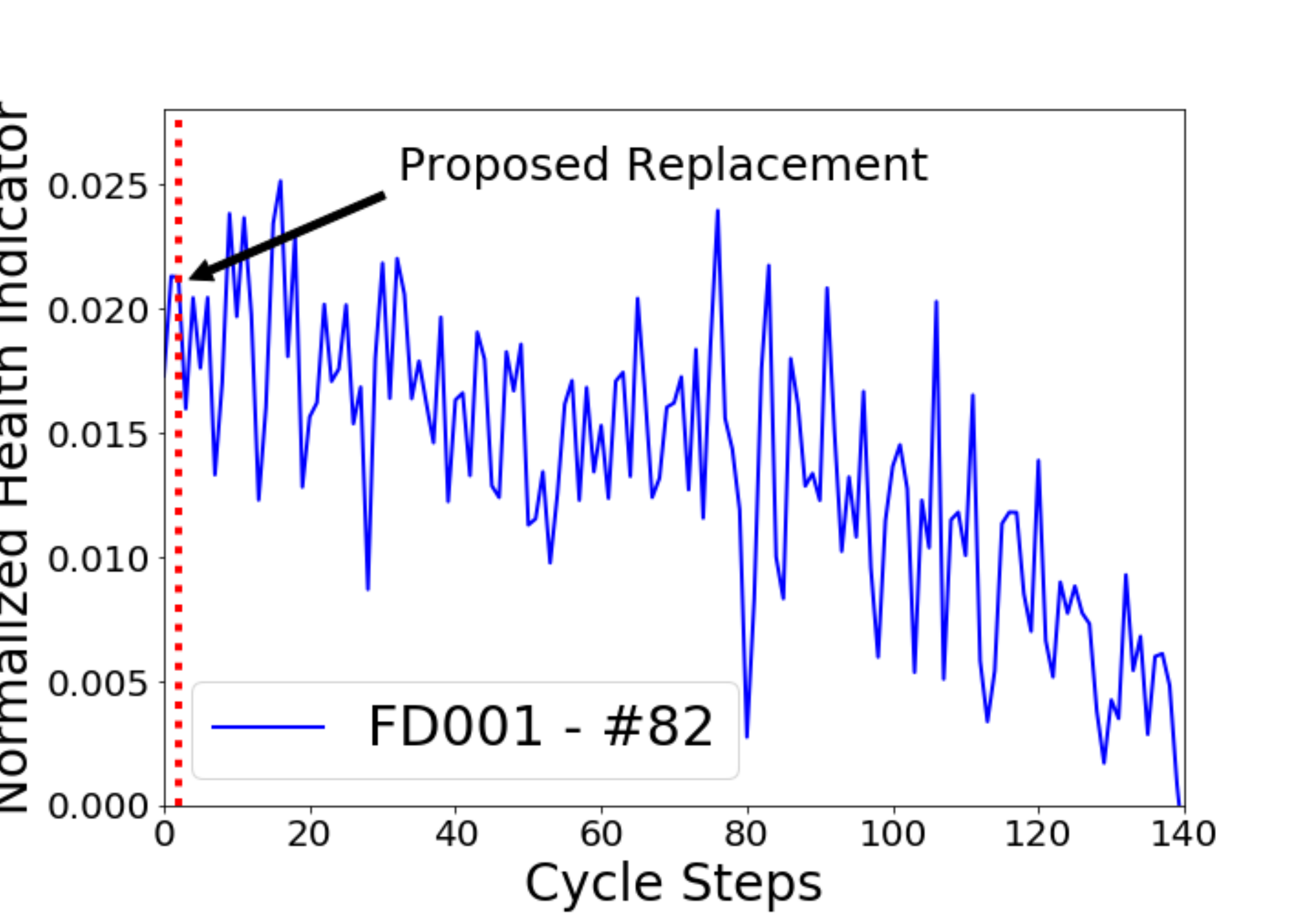}
		\caption{\label{fig:test_fd001_82}Engine82 - (Validation)}
	\end{subfigure}	
	~
	\begin{subfigure}[t]{0.45\linewidth}
		\centering
		\includegraphics[trim={20 5 56 40},clip,width=1\linewidth]{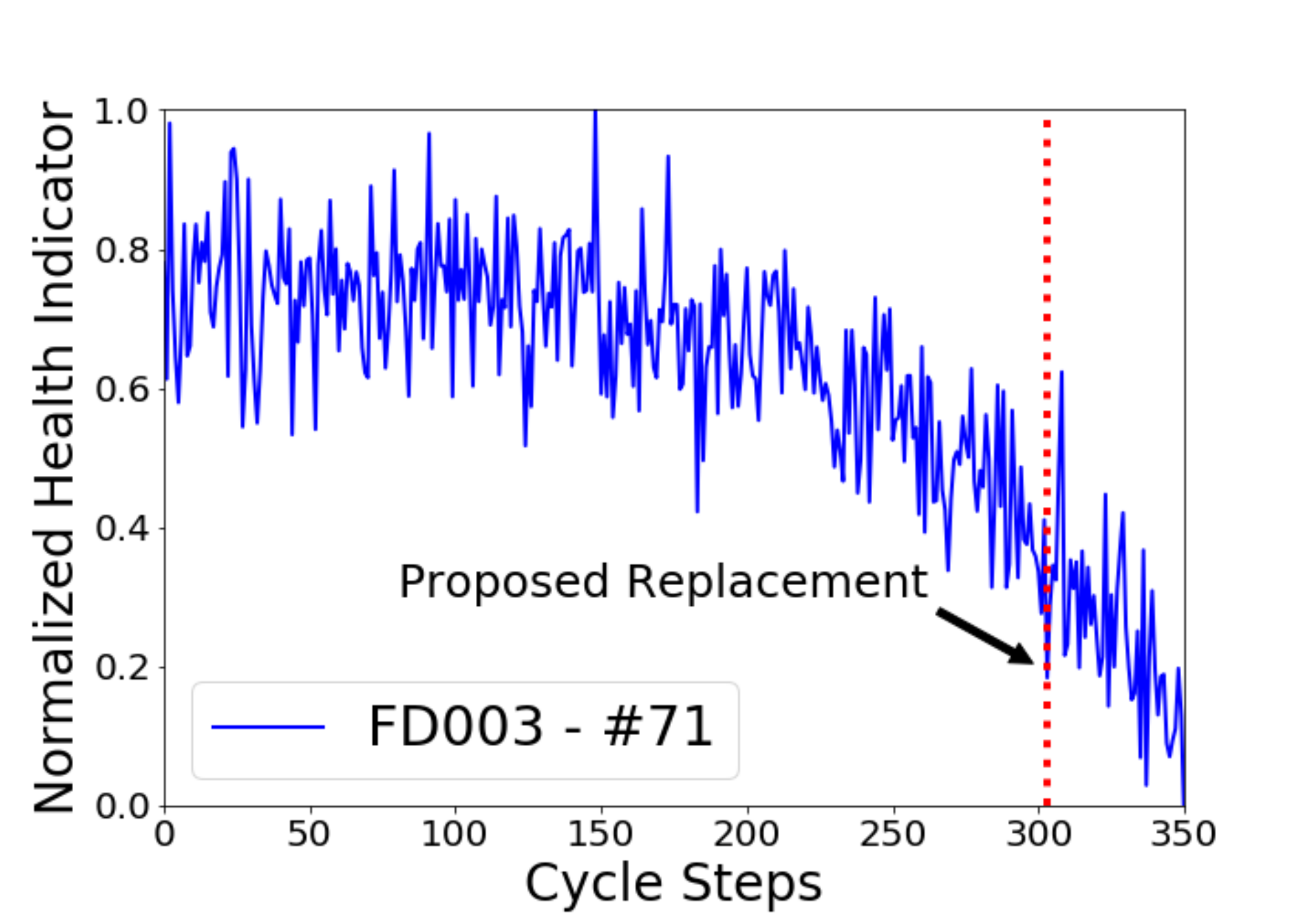}
		\caption{\label{fig:test_fd003_71}Engine71 - (Validation)}
	\end{subfigure}		
	\caption{Example of DRL Agent Proposed Replacement points}
	\label{fig:prediction_results}
	\vspace{-2mm}	
\end{figure}

Part of the validation process involves random selection of Engine Health Indicators from both the training and test dataset with results presented in Figure~\ref{fig:prediction_results}. The DRL agent is able to propose a suitable replacement policy, even when the starting point indicates steep downward trend approaching imminent equipment failure, see Figure~\ref{fig:test_fd001_82}. This is due to the failure-to-failure penalty constraint $\mathcal{R}_{\text{Pen}}$ in ~(\ref{eqn:reward_space}). Likewise, the DRL agent demonstrated encouragingly consistent recommended actions, on different engine and dataset. The proposed replacement points are highlighted in Figure~\ref{fig:prediction_results}. For comparison, cross-validation on FD001 reported a median of 0.175$\pm$0.02; FD003 reported median of 0.174$\pm$0.02.

%The authors noted that the criticality of equipments will differ and a configurable threshold-based health-indicator may be preferable by industrial practitioners. In summary, the PDDQN-PN suggests that the DRL agent demonstrated learn-ability with results that are highly comparable to reported values in Figure~\ref{fig:prediction_results}. 
\vspace{-2mm}

%\subsection{Future Work}
%Future study would include exploring data pre-processing techniques, such as clustering or data augmentation~\cite{jayasinghe2018temporal}, could be considered for improved data pattern extractions when tackling complex or sparse data set. 

\vspace{-2mm}
%==================
\section{Conclusion}
\label{sec:conclusion}
%==================
As industries worldwide journey towards the Industry 4.0 vision to boost manufacturing productivity, modern equipments become increasingly complex to perform maintenance on. The result is an increased customer demand for accurate, interpretable and actionable insights from predictive maintenance tools. In this paper, we have introduced an approach to provide actionable recommendation, based on an equipment's health state. We have formulated the maximization of equipment uptime as a function of multiple input sensor data and model the derived equipment health states with state-action pair. The optimal states are observed in a sparsely-dense configuration and is challenging to solve with existing approaches. Then, we have proposed a model-free-based Deep Reinforcement Learning algorithm which can rapidly learn an optimal maintenance decision policy, for example within 2000 time-steps. The experimental results have shown consistent maintenance recommendations across similar equipment, despite different initial health state. Future work could include extending current work to other equipment failure dataset and benchmark against an actual equipment maintenance policy schedule.

%A latent demand for  of the complex machine sensor data interpretation for predictive maintenance purposes has been assumed. Such data-driven technology should provide actionable yet explainable recommendations, to justify the costly replacement action, and is an important step to reducing unplanned equipment downtime. Unlike traditional black-box regression models, this work describes an approach to provide actionable recommendation, based on an equipment's health status. To the best of the author's knowledge, this is the first work which attempts to formulate the health state of an equipment with actionable outputs. In this work, a model-free algorithm is proposed to tackle the sparse reward (i.e action) problem with fairly consistent recommendations across datasets and similar equipment. For our approach to be more amenable to practitioners, a threshold-based concept was also tested with results that are comparable to originally proposed non-threshold based approach. Future work will include extending current work to more complex dataset, such as FD002 and FD004, propose alternative reinforcement learning based approaches and benchmark against an actual equipment maintenance policy schedule. 
\vspace{-1mm}
\section*{Acknowledgement}
The work was supported in part by Singapore NRF National Satellite of Excellence, Design Science and Technology for Secure Critical Infrastructure NSoE DeST-SCI2019-0007, A*STAR-NTU-SUTD Joint Research Grant Call on Artificial Intelligence for the Future of Manufacturing RGANS1906, WASP/NTU M4082187 (4080), Singapore MOE Tier 1 2017-T1-002-007 RG122/17, MOE Tier 2 MOE2014-T2-2-015 ARC4/15, Singapore NRF2015-NRF-ISF001-2277, and Singapore EMA Energy Resilience NRF2017EWT-EP003-041.
%\vspace{-2mm}
\bibliographystyle{IEEEtran}
\bibliography{Database}

%\]
\end{document}